\documentclass[letterpaper, 10 pt, conference]{ieeeconf}  

\IEEEoverridecommandlockouts                              

\usepackage{multirow}
\usepackage{svg}

\usepackage{graphicx} 
\usepackage{amsmath,amssymb,amsfonts}
\usepackage{hyperref}

\usepackage{cite}

\usepackage{subcaption}

\usepackage{booktabs}
\usepackage{color}
\newcommand*\Update{\color{black}}
\newcommand*\ToDo{\color{black}}
\newcommand*\UpdateFinal{\color{black}}
\newcommand*\Done{\color{black}}

\title{\LARGE \bf
ManeuverNet: A Soft Actor-Critic Framework for Precise Maneuvering of Double-Ackermann-Steering Robots with Optimized Reward Functions
}

\author{ Kohio Deflesselle$^{1*}$, Mélodie Daniel$^{1*}$, Aly Magassouba$^{2}$, Miguel Aranda$^{3}$ and Olivier Ly$^{1}$
\thanks{*These authors contributed equally.}
\thanks{$^{1}$Univ. Bordeaux, CNRS, Bordeaux INP, LaBRI, UMR 5800, F-33400 Talence, France. $^{2}$School of Computer Science, University of Nottingham,
UK. $^{3}$Instituto de Investigación en Ingeniería de Aragón (I3A), Universidad de Zaragoza, 50018 Zaragoza, Spain.  Author’s Accepted Manuscript. Released under the Creative Commons license: Attribution 4.0 International (CC BY 4.0). Corresponding author: Mélodie Daniel, e-mail: \texttt{melodie.daniel@u-bordeaux.fr.}}}%

\begin{document}

\maketitle
\thispagestyle{empty}
\pagestyle{empty}


\begin{abstract}    
Autonomous control of double-Ackermann-steering robots is essential in agricultural applications, where robots must execute precise and complex maneuvers within a limited space. \Update Classical methods, such as the Timed Elastic Band (TEB) planner, can address this problem, but they rely on parameter tuning, making them highly sensitive to changes in robot configuration or environment and impractical to deploy without constant recalibration. At the same time, end-to-end deep reinforcement learning (DRL) methods \Done often fail due to unsuitable reward functions for non-holonomic constraints, resulting in sub-optimal policies and poor generalization. To address these challenges, this paper presents ManeuverNet, a DRL framework tailored for double-Ackermann systems, combining Soft Actor-Critic with CrossQ. Furthermore, ManeuverNet introduces four specifically designed reward functions to support maneuver learning. Unlike prior work, ManeuverNet does not depend on expert data or handcrafted guidance. \Update We extensively evaluate ManeuverNet against both state-of-the-art DRL baselines and the TEB planner. Experimental results demonstrate that our framework substantially improves maneuverability and success rates, achieving more than a 40\% gain over DRL baselines. Moreover, ManeuverNet effectively mitigates the strong parameter sensitivity observed in the TEB planner. In real-world trials, ManeuverNet achieved up to a 90\% increase in maneuvering trajectory efficiency, highlighting its robustness and practical applicability.
 \Done



\end{abstract}

\section{Introduction}
 
\begin{figure} [t]
    \centering
    \captionsetup{font=scriptsize}
    \includegraphics[width=\linewidth]{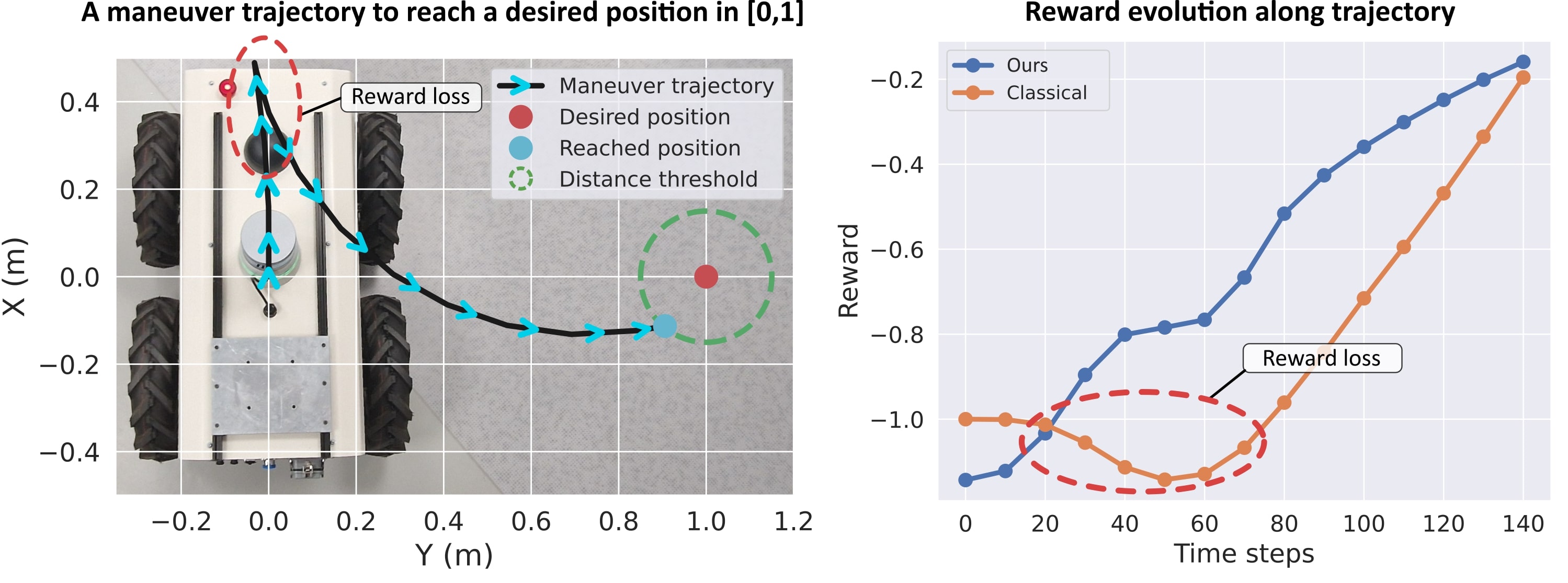}\vspace{-1pt}
    \caption{Maneuver handling (left figure) with 4WS robots in DRL is challenging because it requires current reward loss (circled in red in the right figure), making classical approaches sub-optimal.} \vspace{-0.7cm}
    \label{fig:eye-catch}
\end{figure}

Recent advances in deep reinforcement learning (DRL) for robotics have mainly focused on navigation and locomotion tasks for holonomic robots such as omni-wheeled~\cite{Mehmood2021ITC}, quadruped~\cite{Zhang2024AS}, or biped robots~\cite{GaspardIROS2024}. By holonomic robots, we refer to the ones that can move freely in all directions~\cite{Moreno2016Sensors}. In contrast, Ackermann-steering robots, like cars, represent a restrictive type of non-holonomic robots~\cite{Zhao2013AC}. These robots cannot rotate without moving forward or backward, which reduces their number of degrees of freedom (DOF)~\cite{Siegwart2005}. Such constraints significantly limit their maneuverability and impose additional challenges for control~\cite{SamsonTRO}. In this work, we specifically focus on controlling a mobile platform with four-wheel steering (4WS), and a double-Ackermann-steering mechanism (see Fig.~\ref{fig:eye-catch}). Controlling such robots presents even more complex challenges compared to single-Ackermann-steering systems. The need to coordinate both front and rear steering introduces additional non-holonomic constraints, making precise maneuvering, such as reversing or parallel alignment, even more difficult~\cite{Hulttinen2020}. Despite these challenges, double-Ackermann-steering mobile robots (DASMRs) offer several advantages: improved maneuverability, stability, and energy efficiency~\cite{Yu2010TRO}. These characteristics make them particularly suitable for applications in agriculture, autonomous navigation, and uneven terrain, where robustness and cost-effectiveness are essential~\cite{Deremetz2017ECMR, Thuilot2009}. Enabling these robots to perform precise maneuvers significantly enhances their practical use and autonomy in real-world environments, such as parking in tight spaces, or \Update positioning within recharging areas. \Done 

\Update Classical trajectory optimization and control methods, such as the widely used Timed Elastic Band (TEB) planner~\cite{TEBref}, have been applied to Ackermann and double-Ackermann robots. While these methods can be effective in principle, they require tuning multiple parameters, making them highly sensitive to small variations in robot dynamics (e.g., tire pressure, load distribution). In practice, this limits their robustness and complicates deployment in real-world agricultural environments~\cite{TEB_parameters_limitation}. Therefore, DRL approaches represent a promising alternative. However, the maneuvering constraints of double-Ackermann-steering robots pose a major challenge for DRL agents~\cite{Lazzaroni2022APPLEPIES}. \Done In many scenarios, reaching a desired position requires the robot to perform a complex maneuver, such as initially moving away from the goal to later approach it with the correct orientation. Unfortunately, classic DRL approaches based on reward functions such as the Euclidean distance fail in such scenarios because they penalize the robot for increasing its distance from the desired position, resulting in a reward loss (cf. Fig.~\ref{fig:eye-catch}). As a result, the maneuvers needed to achieve successful positioning are often discouraged. This can lead to sub-optimal policies~\cite{Mihir2021}, where the agent becomes stuck near the goal in an incorrect orientation, unable to complete the task without temporarily sacrificing reward. Achieving such maneuvers can also be challenging when controlled through teleoperation, as even human operators often struggle with precise navigation~\cite{Datar2024IROS}. Therefore, designing specific DRL frameworks that accommodate such constraints is essential for successful learning in DASMRs.

In this paper, we address the control of DASMRs using DRL in a fully model-free setting without relying on expert demonstrations or handcrafted guidance. Our goal is to develop a \Update DRL framework \Done that enables robust and generalizable maneuver learning despite the inherent kinematic constraints of DASMRs. To that end, we propose ManeuverNet, a novel Soft Actor-Critic (SAC) framework for DASMR with optimized reward functions designed specifically to perform precise maneuvers. The contributions of this article can be summarized as follows.

\textbf{- End-to-End DRL Framework:} We propose a fully model-free, end-to-end DRL framework designed to teach DASMRs precise maneuvering without relying on expert data, predefined trajectories, or handcrafted guidance. ManeuverNet ensures robust learning across a variety of environments, and leverages the SAC~\cite{Haarnoja2018SAC} algorithm enhanced with CrossQ~\cite{Bhatt2024CrossQ} for improved sample efficiency and stability during training.

\textbf{- Comprehensive Study of Reward Functions:} We conduct an in-depth study of reward functions for DASMRs, reviewing state-of-the-art reward functions and presenting four novel reward functions for enabling precise maneuvers in non-holonomic robots.

\textbf{- Experimental Validation:} 
We conduct extensive validation of ManeuverNet against a range of DRL and analytical baselines, demonstrating its superior efficiency and robustness. Furthermore, we demonstrate zero-shot transfer capabilities, with the robot consistently performing well in real-world environments across diverse terrains, without any fine-tuning or domain-specific adaptation.


\section{Related Work}   


The control of non-holonomic mobile robots, including those with double-Ackermann-steering mechanisms, has traditionally been tackled using classical control techniques~\cite{Thuilot2009, Hulttinen2020, double-ackermann-2}. \Update Among these, the TEB planner~\cite{TEBref} is widely used to generate smooth trajectories and perform local obstacle avoidance. However, TEB is highly sensitive to parameter tuning, and small changes in robot dynamics, payload, or tire pressure often necessitate full recalibration~\cite{TEB_parameters_limitation}. Its local obstacle handling, based on RANSAC polygon approximations, can also be overly conservative, sometimes causing the robot to halt or oscillate between forward and backward movements rather than progressing toward the goal. \Done 

In recent years, DRL has shown promise for robotic control tasks, but its application has predominantly targeted holonomic systems~\cite{Mehmood2021ITC}. These approaches typically use reward functions $\mathcal{R}_{\text{Euclid}}$ based on Euclidean distance to the goal. While such rewards perform well in holonomic systems, they are not well suited for non-holonomic systems~\cite{Siegwart2005}.

Among non-holonomic platforms, differential-drive robots have been the most studied in DRL research~\cite{ZhangWTYWS25, SOUALHI2025RAS}. Although these robots are non-holonomic, they can rotate in place~\cite{Siegwart2005}, which helps reduce some of the challenges of maneuvering. To guide these robots more effectively, the exponential reward function $\mathcal{R}_\text{Exp}$ was introduced, combining both the Euclidean distance and heading error~\cite{SOUALHI2025RAS}. However, this reward function is less effective for Ackermann-steering robots, as they cannot rotate in place and often need to move away from the goal temporarily in order to align correctly~\cite{Yu2010TRO}. Classic reward functions penalize such maneuvers, resulting in sub-optimal policies. Furthermore, small heading misalignments when near the target can cause significant reward losses, which further contributes to the learned policy being sub-optimal.

Some works have proposed DRL solutions specifically for single-Ackermann-steering robots. A notable example is the FastRLap framework~\cite{Stachowicz2023CoRL}, which trains an agent to follow a predefined racing trajectory. Its reward function $\mathcal{R}_{\text{FastRLap}}$ encourages fast forward motion along the track and handles complex situations using a finite-state machine and expert demonstrations. While effective for high-speed path-following, FastRLap does not directly address the challenge of maneuvering in constrained environments. Furthermore, its reliance on handcrafted guidance and expert input limits its adaptability and generalization across tasks and robot platforms~\cite{DanielRAL2024, Stachowicz2023CoRL}.
Other studies have addressed Ackermann-steering car parking scenarios using reward functions that combine distance and angle error terms~\cite{Lazzaroni2022APPLEPIES, Junzuo2021IOP}. While these reward functions $\mathcal{R}_{\text{Car}}$ effectively address alignment maneuvers when the car is initially perpendicular to the desired position, they fall short when dealing with more complex maneuvers. Specifically, they do not account for situations in which the robot must rotate and move along different axes to reach the goal. 

To overcome local reward minima and sub-optimal policy convergence in DRL more broadly, two major strategies have been explored. The first is curriculum learning, which gradually increases task complexity to facilitate learning~\cite{Honghu2022AS}. This often involves breaking down a complex goal into intermediate waypoints to help guide the agent. However, all methods that rely on guided training, whether via curriculum learning, imitation learning, or supervised learning, tend to be highly task and environment-specific~\cite{DanielRAL2024, Stachowicz2023CoRL}. The second is Hindsight Experience Replay (HER)~\cite{Andrychowicz2017HER}, which reframes failed trajectories as successful by redefining the goal retroactively. HER uses sparse reward functions $\mathcal{R}_{\text{HER}}$, avoiding penalizing exploratory behaviors that are essential for reaching the goal. HER has proven to be a powerful general-purpose strategy, but it is not tailored to the specific challenges of double-Ackermann-steering control. 

Despite growing interest, to the best of our knowledge, no end-to-end DRL framework has been proposed specifically for DASMR, which are more difficult to maneuver. This highlights a critical gap that our work aims to address.

\section{Problem Statement} \label{PS}
\begin{figure}[!t]
  \centering
  \captionsetup{font=scriptsize}
  \includegraphics[width=0.7\linewidth]{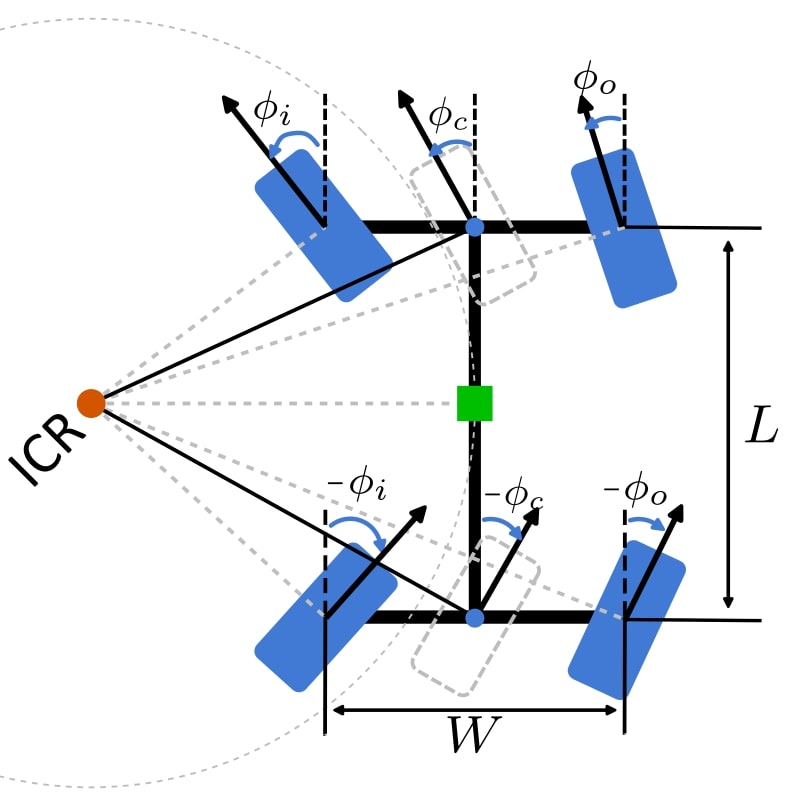}\vspace{-1pt}
  \caption{DASMR rotating around an instantaneous center of rotation (ICR).} \vspace{-0.5cm}
  \label{fig:ackermann-geom}
\end{figure}

We consider a DASMR, controlled to make a maneuver to reach a desired 2D position $\boldsymbol{X_d} = (x_d,y_d)$. The DASMR is a non-holonomic platform. Its configuration includes $(\boldsymbol{X_c}, \theta_c)$, where $\boldsymbol{X_c}=(x_c,y_c)$ denotes the center position and $\theta_c$ the orientation of its longitudinal axis (cf. Fig.~\ref{fig:ackermann-geom}). However, only two DOF (forward motion and change of orientation) can be directly controlled, and the  
DASMR cannot rotate without moving forward or backward. In this paper, the double-Ackermann configuration is used in a symmetric, negative 4WS setup, where the front and rear wheel steering angles are mirrored, i.e., they are equal in magnitude but opposite in direction relative to the chassis-fixed frame as shown in Fig.~\ref{fig:ackermann-geom}. Although ManeuverNet can be applied to any kind of DASMR, in this study, we consider large and heavy ($>$ 50 kg) robots, which are generally used in agricultural settings. The robot is initially positioned at the center of a 8-meter side square workspace and can move freely in this environment. The objective is to control the spinning velocity $\omega$ and steering angle $\phi$ of the four robot wheels to reach $\boldsymbol{X_d}$. 

A key challenge is enabling the robot to generate feasible maneuvers without relying on prior expert knowledge or pre-defined trajectories. In addition, we assume a model-free setting in which the robot's dynamics are unknown. The robot's wheel spinning acceleration and steering speeds are limited and empirically determined to ensure stability during motion. We train a DRL agent to generate control commands that respect these physical constraints while achieving the desired behavior. This agent interacts with a simulated environment to learn a policy $\pi$ that maximizes cumulative rewards over time. The agent is designed according to the Markov Decision Process (MDP) formalism~\cite{SuttonMIT2018}. An MDP is defined as the tuple $(\mathcal{S}, \mathcal{A}, \mathcal{P}, \mathcal{R})$, where $\mathcal{S}$ is the state space, $\mathcal{A}$ is the action space, $\mathcal{P}$ is the state transition function, and $\mathcal{R}$ is the reward function. At each discrete time step $t$, the agent observes a state $s_t \in \mathcal{S}$, selects an action $a_t \in \mathcal{A}$ according to its current policy, and transitions to a new state $s_{t+1}$ based on $\mathcal{P}$. It then receives a reward $r_t = R(s_t, a_t)$ as feedback. The policy $\pi : \mathcal{S} \rightarrow \mathcal{A}$ is deterministic and maps each state to a specific action. While the distance between the robot and the target point is defined by $d$, the success of reaching $\boldsymbol{X_d}$ is determined by a distance threshold $d_{th}$. Similar to many DRL frameworks~\cite{Zhao2020SimtoReal}, ManeuverNet is trained in a simulator. 




\vspace{-0.3cm} \section{Method}
In our framework, a DRL agent controls $\omega$ and $\phi$. At the beginning of each training episode, the robot is initialized in the center of a square obstacle-free workspace, and must reach $\boldsymbol{X_d}$. 

\subsection{RL Background}
In robotics, actor-critic algorithms have been successfully used in several control tasks~\cite{DanielRAL2024, GaspardICRA2024, GaspardIROS2024}. Actor-critic algorithms rely on the interaction between two dense neural networks (DNNs): an actor and a critic~\cite{DanielRAL2024}. The actor $\mu$, also called the policy network, selects an action $a_t$ based on the current state $s_t$, following $a_t = \mu(s_t)$. The critic $Q$, also called the Q-network, estimates the expected return of the state-action pair $(s_t, a_t)$ by computing the Q-value $Q^\pi (s_t, a_t)$. The critic is updated using the temporal difference learning and the Bellman equation~\cite{SuttonMIT2018}, with $Q^\pi_t (s_t, a_t) = r_t + \gamma \mathbb{E} [Q^\pi_{t+1} (s_{t+1}, a_{t+1})]$. The actor is updated by maximizing the expected Q-value.

\begin{figure*}[!ht] 
    \centering
    \captionsetup{font=scriptsize}
    \begin{subfigure}[t]{0.24\textwidth}
        \centering
        \captionsetup{font=scriptsize}
        \includegraphics[width=\textwidth]{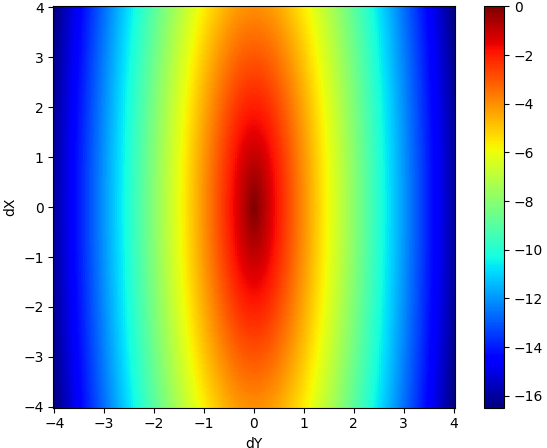}
        \caption{\scriptsize $\mathcal{R}_\text{ES}$ shape with $c = 4.0$}
        \label{sub:ellipse-shape}
    \end{subfigure}\hfill%
    \begin{subfigure}[t]{0.24\textwidth}
        \centering
        \captionsetup{font=scriptsize}
        \includegraphics[width=\textwidth]{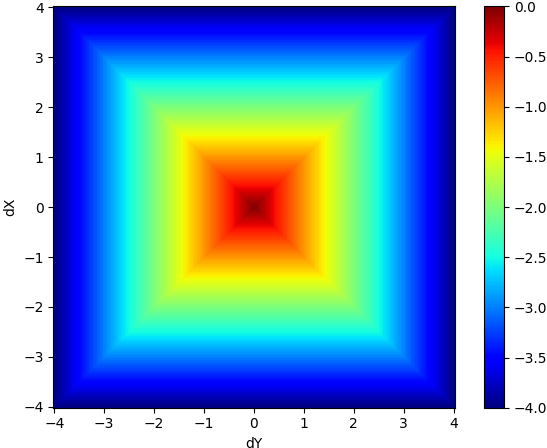}
        \caption{\scriptsize $\mathcal{R}_\text{Ch}$ shape}
        \label{sub:tcheby-shape}
    \end{subfigure}\hfill%
    \begin{subfigure}[t]{0.24\textwidth}
        \centering
        \captionsetup{font=scriptsize}
        \includegraphics[width=\textwidth]{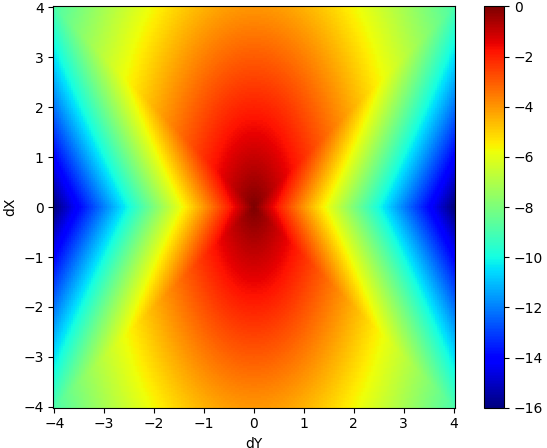}
        \caption{\scriptsize $\mathcal{R}_\text{HS}$ shape with $c = 2.0$}
        \label{sub:hour-shape}
    \end{subfigure}\hfill%
    \begin{subfigure}[t]{0.24\textwidth}
        \centering
        \captionsetup{font=scriptsize}
        \includegraphics[width=\textwidth]{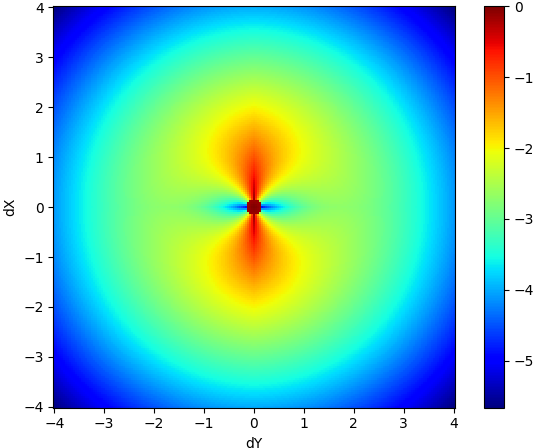}
        \caption{\scriptsize $\mathcal{R}_\text{Cl}$ shape with $c = 3.0$}
        \label{sub:clover-shape}
    \end{subfigure}\\%
    \vspace{0.1cm}%
    \begin{subfigure}[t]{0.24\textwidth}
        \centering
        \captionsetup{font=scriptsize}
        \includegraphics[width=\textwidth]{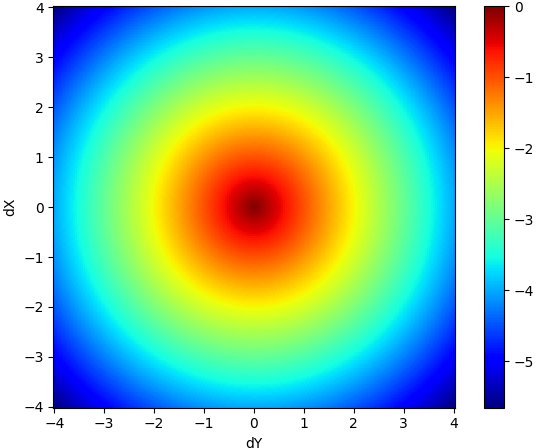}
        \caption{\scriptsize $\mathcal{R}_\text{Euclid}$ shape}
        \label{sub:euclid-shape}
    \end{subfigure}\hfill%
    \begin{subfigure}[t]{0.24\textwidth}
        \centering
        \captionsetup{font=scriptsize}
        \includegraphics[width=\textwidth]{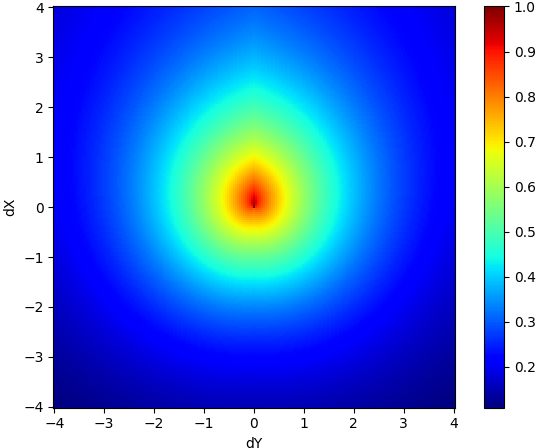}
        \caption{\scriptsize \UpdateFinal $\mathcal{R}_\text{Exp}$ shape with $K = 0.5$, $\lambda_d = 0.8$ and $\lambda_{\theta} = 0.2$~\cite{SOUALHI2025RAS} \Done}
        \label{sub:dnh-shape}
    \end{subfigure}\hfill%
    \begin{subfigure}[t]{0.24\textwidth}
        \centering
        \captionsetup{font=scriptsize}
        \includegraphics[width=\textwidth]{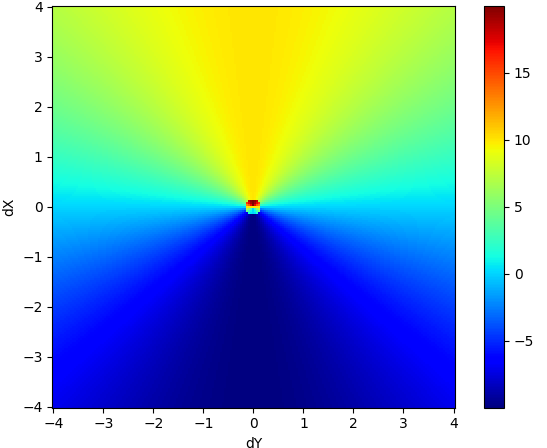}
        \caption{\scriptsize $\mathcal{R}_\text{FastRLap}$ shape with $V$ fixed at $[10, 0]$ m/s~\cite{Stachowicz2023CoRL}}
        \label{sub:fast-shape}
    \end{subfigure}\hfill%
    \begin{subfigure}[t]{0.24\textwidth}
        \centering
        \captionsetup{font=scriptsize}
        \includegraphics[width=\textwidth]{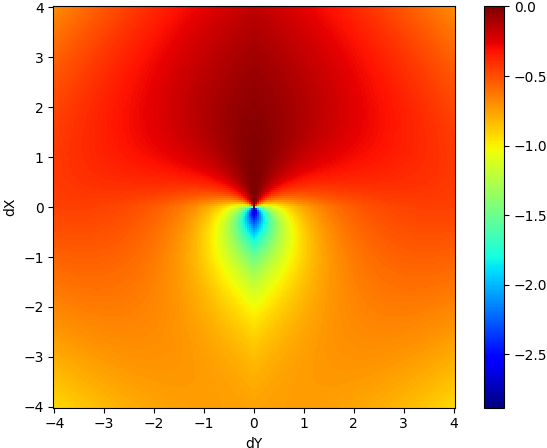}
        \caption{$\mathcal{R}_\text{Car}$ shape with $c_1 = 0.01$ and $c_2 = 1.5$~\cite{Lazzaroni2022APPLEPIES}}
        \label{sub:sparse-shape}
    \end{subfigure}\vspace{-4pt}
    \caption{Comparison between the shape of our reward functions (top) vs. classic state-of-the-art reward functions (bottom). X is the longitudinal axis, Y is the lateral axis, and the robot's center position is at the origin. Each point on the heatmaps represents the reward for a $\boldsymbol{X_d}$ in the same position. Thus, the robot reaches $\boldsymbol{X_d}$ when $\|\boldsymbol{X_d} \| < d_\text{th}$.}
    \label{fig:reward-shapes} \vspace{-0.4cm}
\end{figure*}

\subsection{DRL Algorithms}
SAC, introduced in~\cite{Haarnoja2018SAC}, optimizes both cumulative rewards and policy entropy to encourage exploration and improve stability. Additionally, it uses two critic networks, a technique from Twin Delayed Deep Deterministic Policy Gradient (TD3)~\cite{Fujimoto2018TD3}, to mitigate Q-value overestimation bias. 

Despite these improvements, SAC is still computationally expensive. To improve learning efficiency, the CrossQ algorithm~\cite{Bhatt2024CrossQ} has been recently proposed as an SAC overlay that eliminates target networks~\cite{Lillicrap2015DDPG}. CrossQ introduces batch normalization layers in the DNN of the actor and the critic, and employs wider critic layers. Experimental results in~\cite{GaspardICRA2024, Bhatt2024CrossQ} showed that CrossQ outperforms existing actor-critic algorithms while significantly reducing computational times and increasing sample efficiency, making it an interesting solution for robotics applications. Therefore, our DRL agent leverages the SAC algorithm, enhanced with the CrossQ overlay.



\subsection{State Space}
As described in Section~\ref{PS}, the robot's center position is denoted by $\boldsymbol{X_c}$ and its orientation by the yaw angle $\theta_c$. The spinning velocities of the left and right wheels are $\omega_l$ and $\omega_r$. The steering angles of the left and right wheels are $\phi_l$ and $\phi_r$, and the corresponding steering velocities $\dot{\phi}_l$ and $\dot{\phi}_r$. The robot's center linear velocity is $\boldsymbol{V_c} = (\dot x_c, \dot y_c)$, and its angular velocity is $\dot{\theta_c}$. At the time step $t$, we define the DRL agent's current state $\boldsymbol{s_t} \in \mathcal{S}$ as ($\boldsymbol{X_c}$, $\boldsymbol{X_d}$, $\theta_c$, $\omega_l$, $\omega_r$, $\phi_l$, $\phi_r$, $\dot\phi_l$, $\dot\phi_r$, $\boldsymbol{V_c}$, $\dot\theta_c$) $\in \mathbb{R}^{14}$.

\vspace{-0.cm}\subsection{Action Space}
When neglecting steering dynamics and assuming ideal Ackermann-steering, the kinematics of the platform can be approximated by a simplistic bicycle model, reducing front and rear pairs of wheels to a single pair of virtual central wheels~\cite{Hulttinen2020}. Since we consider a negative symmetric 4WS configuration, the rear steering angles mirror the front steering angles. Let us define the spinning velocity and the steering angle of the virtual central wheels as $\omega_{c}$ and $\phi_{c}$, respectively. At each time step $t$, the DRL agent outputs an action $\boldsymbol{a_t} = (\omega_c, \phi_c) \in [-1, 1]$, representing normalized commands for the wheel spinning velocity and the steering angle. These values are then scaled by the vehicle’s respective actuation limits to obtain the actual control inputs. Following the double-Ackermann-steering geometry, the spinning velocities and steering angles of the left and right wheels can be calculated from $\omega_{c}$ and $\phi_{c}$. For this, we compute the steering angles of the inner and outer wheels $\phi_{i}$ and $\phi_{o}$, relative to the robot's current instantaneous center of rotation (ICR) shown in Fig.~\ref{fig:ackermann-geom}, as: 
\begin{align}
    \phi_i = \tan^{-1}\frac{2 L \cdot \sin{\phi_{c}}} {2 L \cdot \cos{\phi_{c}} - W \cdot \sin{\phi_{c}}} 
\end{align} \vspace{-0.cm}
\begin{align}
   \text{and }  \phi_o = \tan^{-1}\frac{2 L \cdot \sin{\phi_{c}}} {2 L \cdot \cos{\phi_{c}} + W \cdot \sin{\phi_{c}}},
\end{align}
where $L$ is the wheelbase of the vehicle and $W$ is the track of the vehicle. Similarly, we can compute the spinning velocities of the inner and outer wheels $\omega_{i}$ and $\omega_{o}$ relative to the ICR as:
\begin{align}
    \omega_{i} = \omega_{c} \frac{ \sqrt{[L \cdot \tan (\frac{\pi}{2} - |\phi_{i}|)]^2 + L^2 } }{ \sqrt{[L \cdot \tan (\frac{\pi}{2} - |\phi_{i}|) + \frac{W}{2}]^2 + L^2} } 
\end{align} \vspace{-0.cm}
\begin{align}
    \text{ and } \omega_{o} = \omega_{c} \frac{ \sqrt{[L \cdot \tan(\frac{\pi}{2} - |\phi_{i}|) + W]^2 + L^2 } }{ \sqrt{[L \cdot \tan(\frac{\pi}{2} - |\phi_{i}|) + \frac{W}{2}]^2 + L^2} }.
\end{align}

These inner and outer wheel spinning velocities and steering angles can then be applied to the left and right wheels based on the sign of $\phi_{c}$. If $\phi_c \geq 0$, the left wheel is the inner wheel, so that $\omega_l = \omega_i$ and $\phi_l = \phi_i$, while the right wheel becomes the outer wheel with $\omega_r = \omega_o$ and $\phi_r = \phi_o$. If $\phi_c < 0$, the roles are reversed, with the right wheel becoming the inner wheel.


\subsection{Reward Functions}\label{reward_new_section}

To address the issue of sub-optimal policies for DASMRs, we introduce the reward $\mathcal{R}_\text{HS}$ (cf. Table  \ref{table:rewards}). This reward is designed to handle scenarios where the robot must temporarily deviate from the target to execute a successful maneuver. By prioritizing lateral ($Y$-axis) error over longitudinal ($X$-axis) error, $\mathcal{R}_\text{HS}$ minimizes reward penalties when the robot temporarily moves away from the goal. As shown in Fig.~\ref{sub:hour-shape}, this reward is shaped as an hourglass, which reduces penalization during maneuvers.

With the same objective, we also investigate alternative reward functions: $\mathcal{R}_\text{ES}$, $\mathcal{R}_\text{Ch}$, and $\mathcal{R}_\text{Cl}$, each defined in Table  \ref{table:rewards}. The reward $\mathcal{R}_\text{ES}$ is a variation of the Euclidean distance that scales the error components asymmetrically, forming an ellipse. The reward $\mathcal{R}_\text{Ch}$ is based on the Chebychev distance focusing on the maximum deviation along either axis, while $\mathcal{R}_\text{Cl}$ extends the exponential reward, proposed in \cite{SOUALHI2025RAS}, by incorporating a directional weighting term specifically tuned for DASMR dynamics. This reward penalizes large lateral deviations and misalignments only when the robot is close to the target $\boldsymbol{X_d}$. A comparative visualization of these reward functions alongside conventional formulations is provided in Fig.~\ref{fig:reward-shapes}. \vspace{-0.1 cm}

\begin{table}[!h]
\captionsetup{font=scriptsize}
\caption{DASMR reward functions  where $\Delta x$ and $\Delta y$ are the X and Y components of $\left(\boldsymbol{X_d} -\boldsymbol{X_c}\right)$ and $c$ a weighting parameter.} \vspace{-0.4cm}
\label{table:rewards}
\begin{center}
\begin{tabular}{l|c} 
\toprule
\textbf{Rewards} & \textbf{Formulation} \\
\hline
\\[-1em]
$\mathcal{R}_\text{HS}$ & $-\sqrt{(\Delta x)^2 + \left(c \cdot (\Delta y \pm \max \left\{0, |\Delta y| - |\Delta x| \right\})\right)^2}$ \\
\hline
\\[-1em]
$\mathcal{R}_\text{ES}$ & $-\sqrt{(\Delta x)^2 + (c\Delta y)^2}$\\
\hline
\\[-1em]
$\mathcal{R}_\text{Ch}$ &$-\max \left\{ |\Delta x|, |\Delta y| \right\}$\\
\hline
\\[-1em]
$\mathcal{R}_\text{Cl}$ & 
$\begin{cases} 
\tan^{-1} \left( \frac{\Delta y}{\Delta x} \right) \times c e^{d_\text{th}- d} &\text{ if  $\Delta y > d_\text{th}$ }\\
-d &\text{otherwise} \\
\end{cases}$
\\
\bottomrule 
\end{tabular} \vspace{-0.3cm}
\end{center}
\end{table}

\section{Experimental Results}
\label{sec:result}

\subsection{Environment Setup}

\begin{figure}[!h]

    \captionsetup{font=scriptsize}
    \centering 
    
    \begin{subfigure}[b]{0.75\linewidth}
        \includegraphics[width=\linewidth]{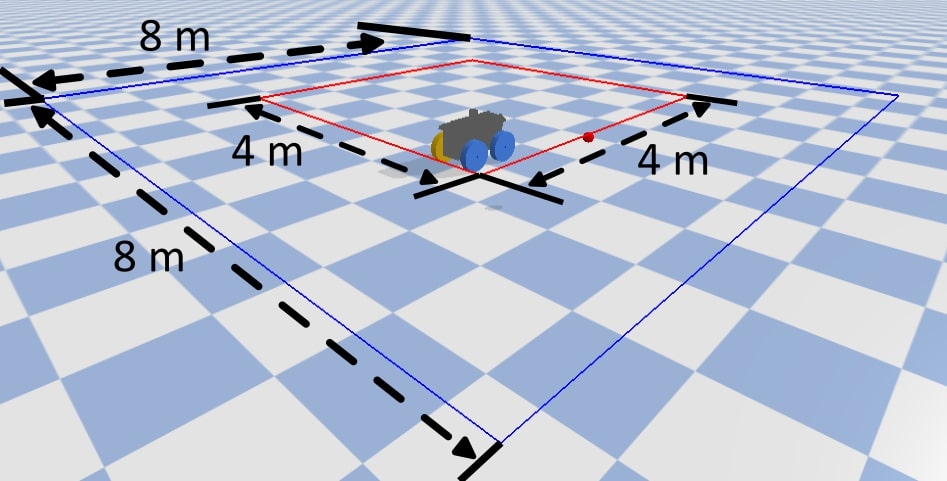}
    \end{subfigure}
    
    \vspace{0.4em}
    
    \begin{subfigure}[b]{0.75\linewidth}
        \includegraphics[width=\linewidth]{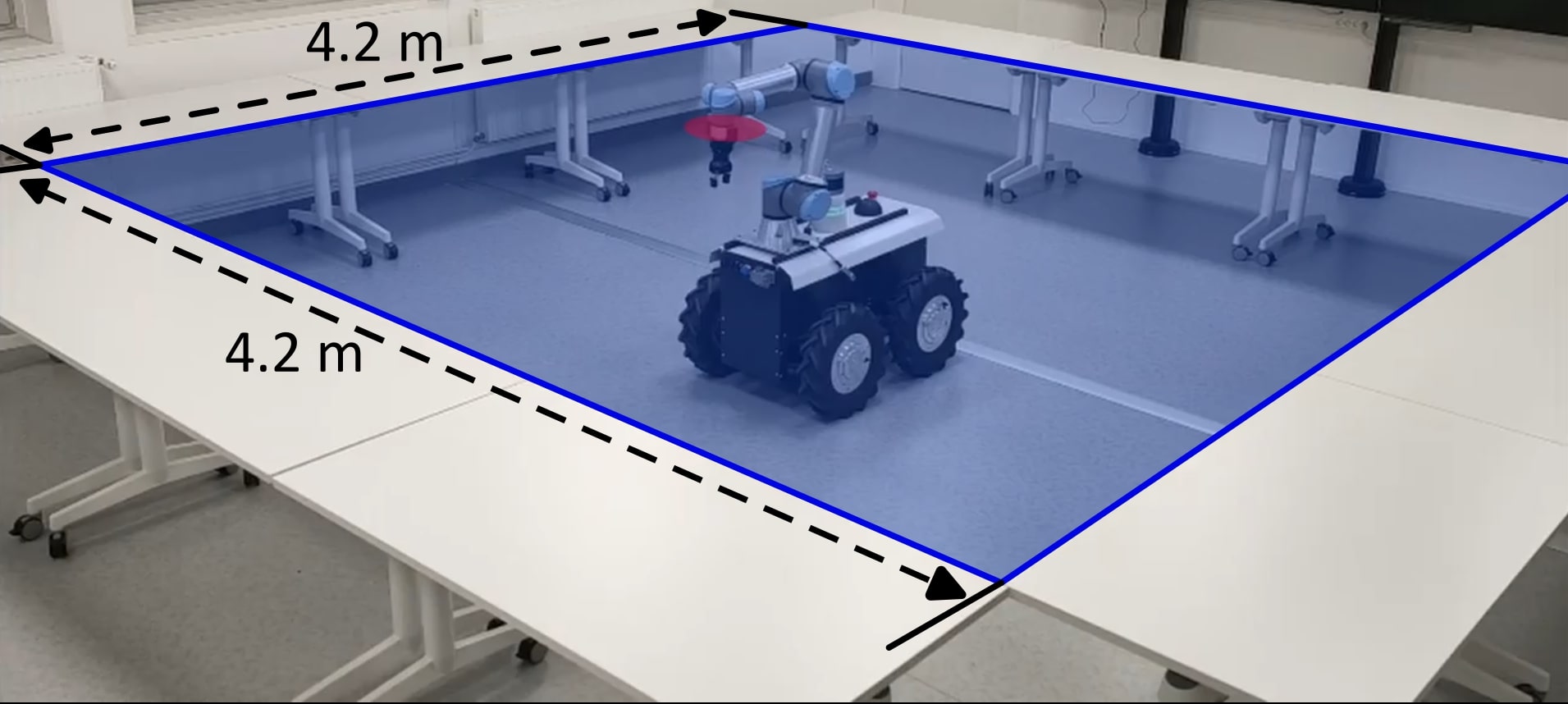}
    \end{subfigure}    
    \vspace{-0.1cm}\caption{Environment setup in simulation and real-world settings. The red square denotes the goal space. The blue square represents the robot's workspace. In real-world settings, both spaces coincide to impose stricter constraints on navigation and positioning.} \vspace{-0.5cm}
    \label{fig:env_spaces}%
    
\end{figure}

We considered the Shadow Runner RR100 EDU rover as our mobile robot platform. The RR100 weighs approximately 100 kg and is 65 cm wide, 90 cm long, and 80 cm high. The DRL agent controlling the robot was trained using the PyBullet simulator, in an environment defined as follows: the robot is \UpdateFinal placed at coordinates $(0,0)$, \Done and its movements are limited to an $8\times8\text{ m}^2$ square workspace. When the environment is reset, the robot is reset to initial configuration, and a new $\boldsymbol{X_d} = (x_d, y_d)$ is sampled from a $4 \times 4$ m\textsuperscript{2} goal space. All values, including spaces and goal coordinates, are expressed in the robot's frame at reset. The rationale for using a small goal space is to maximize the number of $\boldsymbol{X_d}$ requiring complex maneuvers. In real-world settings, the robot's workspace and goal space are constrained to the same $4.2 \times 4.2$ m\textsuperscript{2} square, resulting in a more restrictive setup that better reflects practical operating conditions. \Update Fig.~\ref{fig:env_spaces} presents the environment setup used in both simulation and real-world experiments. \Done


A training episode is considered successfully terminated when the DRL agent reaches $\boldsymbol{X_d}$ within $d_\text{th}$, regardless of orientation. If the robot fails to reach $\boldsymbol{X_d}$ within a time step limit or drives outside the bounds of its workspace, the episode is truncated, and the environment is consequently reset. 
\UpdateFinal We released the code at \url{https://github.com/MelodieDANIEL/4ws_actor_critic_maneuvering}.\Done 



\begin{table}[!tb]
\captionsetup{font=scriptsize}
\caption{SAC and CrossQ parameters used on SBX during training and testing.}\vspace{-0.4 cm} 
\scriptsize
\label{table:hyperparams}
\begin{center}
\scalebox{0.8}{
\begin{tabular}{c|c} 
\toprule
\textbf{Parameter} & \textbf{Value} \\
\hline
Nb. layers & 2 \\
Actor Hidden size & 256 \\
Critic Hidden size & 1024 \\
$\alpha_A = \alpha_C$ & 1e-3 \\
Replay buffer size & 1,000,000 \\
Batch size $N$ & 256 \\ 
$\gamma$ & 0.99 \\ 
Training $d_\text{th}$ & 15 cm \\ 
Training random seed & 9527 \\ 
Total testing time steps & 100,000 \\ 
Nb. testing time steps limit per episode for settings (i), (ii) \& (iii) & 300 \\ 
Nb. testing time steps limit per episodes for setting (iv) & 500 \\ 
Testing random seed & 10, 52, 1234, and 9527 \\  
Testing $d_\text{th}$ & 10, and 15 cm \\ 
\bottomrule
\end{tabular}} \vspace{-0.5 cm}
\end{center}
\end{table}


 \subsection{Training Setup}
The DRL agent was trained in a single, non-vectorized simulation environment for 600,000 time steps, where each episode was limited to 800 time steps before truncation. The agent interacted with its environment at a frequency of 40 Hz, resulting in 40 time steps per second. The training progress was monitored by logging the average episode reward and success rate (SR) every 10 episodes, both computed using a sliding window of 100 episodes. The full training process took approximately 1.5 hours. The training was carried out on a desktop computer equipped with an AMD Ryzen Threadripper PRO 7985WX 64-Cores (AMD Zen 4) CPU, 128 GB of memory, along with an Nvidia RTX 4090 GPU. The DRL algorithms were implemented using the Stable Baselines JAX (SBX) library, leveraging its CrossQ implementation built upon the SAC algorithm. \Update Hyperparameters and network architecture details for both SAC and CrossQ are provided in Table~\ref{table:hyperparams}, and were kept consistent in all experiments unless otherwise specified. To ensure reproducibility, detailed network architecture and algorithm parameters are available in our GitHub repository. \Done

\Update

\subsection{Benchmarking Against Baseline Approaches in Gazebo}
To evaluate the effectiveness and generalizability of ManeuverNet, we conducted experiments in the Gazebo simulator, which offers a more realistic physics engine than PyBullet, thereby helping to assess robustness across different simulators (sim-to-sim gap). ManeuverNet combines the SAC framework with CrossQ, together with the reward $\mathcal{R}_\text{HS}$ specifically designed to encourage maneuvers. We compared ManeuverNet with two representative DRL baselines: (i) a standard DRL agent trained with the SAC framework using a Euclidean distance-based reward commonly used for mobile robot navigation~\cite{DRLStandard}, and (ii) the single-Ackermann framework FastRLap~\cite{Stachowicz2023CoRL} implemented with the SAC parameters reported in the original publication. To the best of our knowledge, FastRLap is the only available DRL framework for Ackermann mobile robots. In addition to these DRL-based methods, we benchmarked against the analytical TEB planner~\cite{TEBref}. This closed-loop online local planner relies on several inputs, including the robot’s current velocity, a local occupancy map, and its global pose estimation, to generate feasible trajectories in real time.

All evaluations were conducted using a different random seed than during training, ensuring exposure to 20 unseen goals. Furthermore, unlike the training setup, the robot was not reset to its initial position after each episode. This forced the agent to handle successive tasks continually, without relying on episodic resets to simplify maneuvering. These evaluation conditions were designed to reflect real-world deployment scenarios and rigorously test the robustness of each method under dynamic and non-ideal circumstances.

Each approach was quantitatively evaluated with the success rate (SR), and the average distance error (AE) and the standard deviation to $\boldsymbol{X_d}$. \ToDo We also evaluated the average Success weighted by (normalized inverse) Path Length (SPL)~\cite{AndersonSPL}. High SPL values indicate trajectories that are not only successful but also closely aligned with the shortest possible path. Achieving a high SPL is particularly challenging for DASMRs, as reaching the target position often requires complex maneuvers deviating from the optimal path. \Done


\Update As shown in Table~\ref{table:sim-to-sim},  although ManeuverNet does not reach the SR of the TEB planner, it consistently surpasses other DRL baselines. It demonstrates superior maneuvering proficiency while achieving shorter, more efficient trajectories, as evidenced by the average SPL metric. Interestingly, if we only focus on the successfully reached targets, ManeuverNet has a better SPL than TEB (0.89). The AE metric reveals that the standard DRL method gets relatively close to the target but it fails to attain precisely the desired position, primarily due to its limited maneuverability. Differently, FastRLap tends to drift away from the target and not maintain proximity, due to its emphasis on forward motion. \Done

\Update
\subsection{Comparative Study of the Reward Functions} 




\begin{figure*}[!h]
    \captionsetup{font=scriptsize}
    \centering
    
    \begin{subfigure}[b]{0.19\textwidth}
        \includegraphics[width=\textwidth]{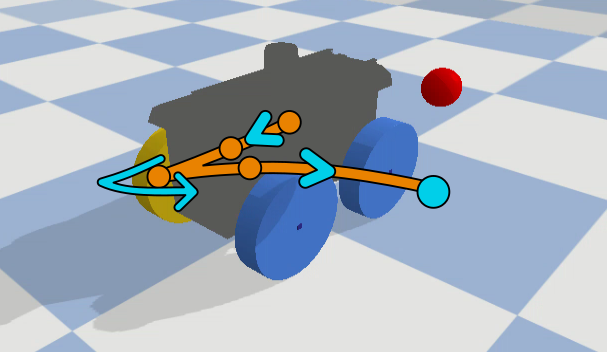}
    \end{subfigure}
    \hfill
    \begin{subfigure}[b]{0.19\textwidth}
        \includegraphics[width=\textwidth]{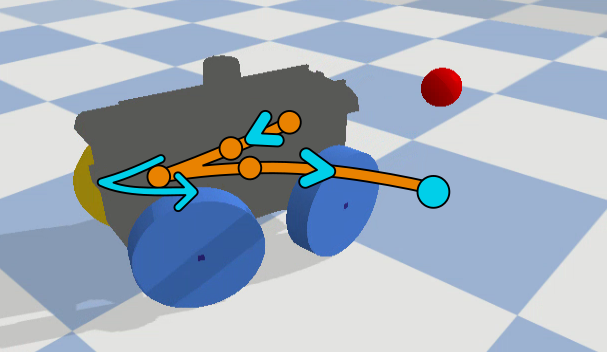}
    \end{subfigure}
    \hfill
    \begin{subfigure}[b]{0.19\textwidth}
        \includegraphics[width=\textwidth]{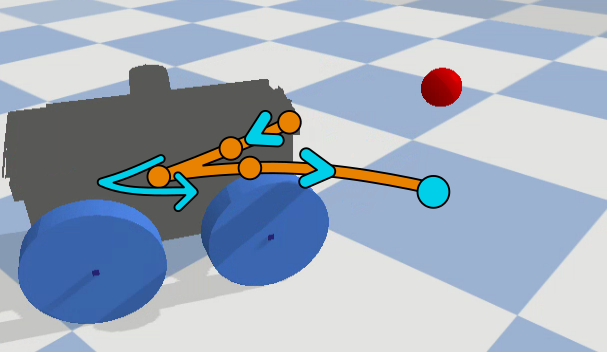}
    \end{subfigure}
    \hfill
    \begin{subfigure}[b]{0.19\textwidth}
        \includegraphics[width=\textwidth]{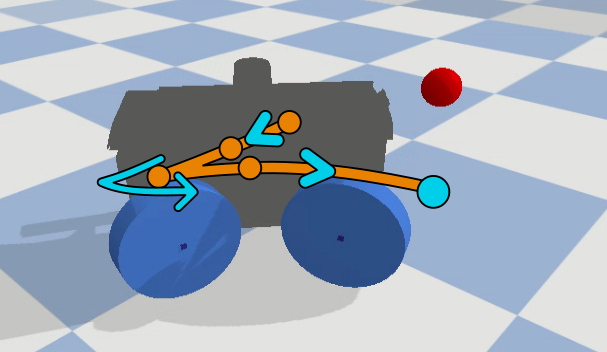}
    \end{subfigure}
    \hfill
    \begin{subfigure}[b]{0.19\textwidth}
        \includegraphics[width=\textwidth]{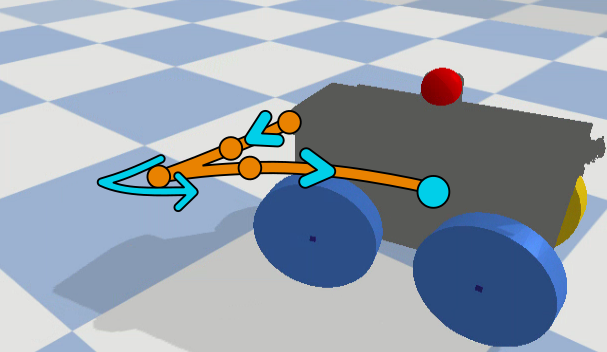}
    \end{subfigure}

    \vspace{0.2cm}

    \begin{subfigure}[b]{0.19\textwidth}
        \includegraphics[width=\textwidth]{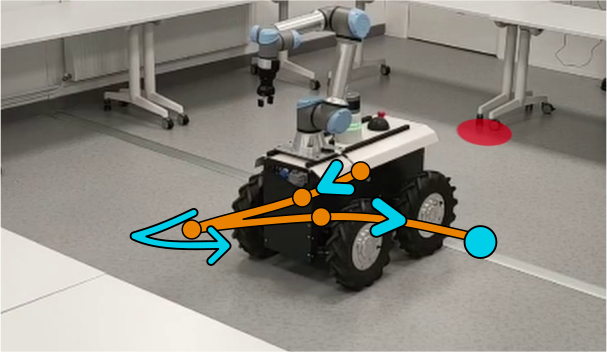}
    \end{subfigure}
    \hfill
    \begin{subfigure}[b]{0.19\textwidth}
        \includegraphics[width=\textwidth]{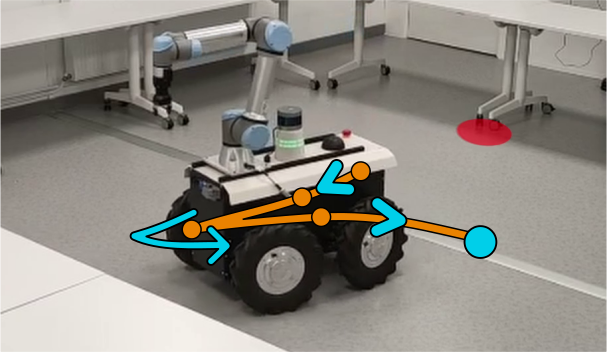}
    \end{subfigure}
    \hfill
    \begin{subfigure}[b]{0.19\textwidth}
        \includegraphics[width=\textwidth]{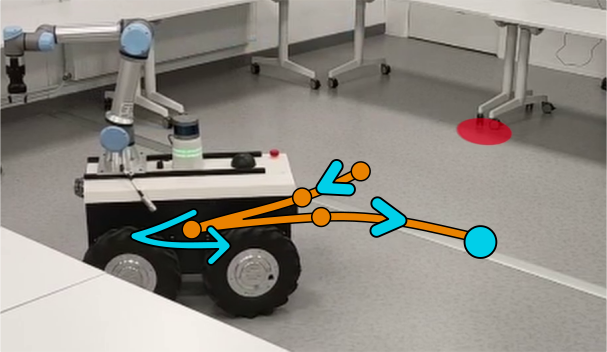}
    \end{subfigure}
    \hfill
    \begin{subfigure}[b]{0.19\textwidth}
        \includegraphics[width=\textwidth]{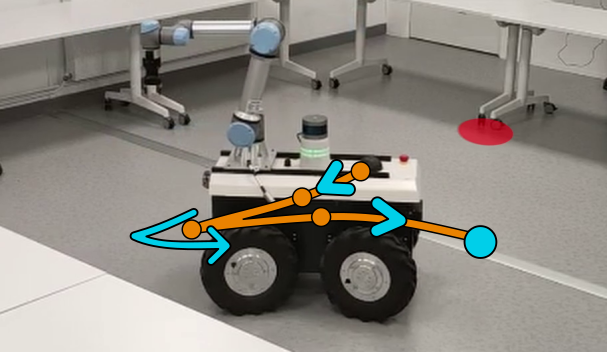}
    \end{subfigure}
    \hfill
    \begin{subfigure}[b]{0.19\textwidth}
        \includegraphics[width=\textwidth]{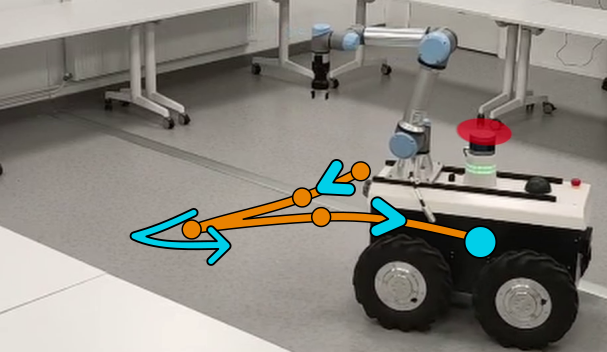}
    \end{subfigure}
    
    \vspace{-0.15cm}
    \dotfill
    \vspace{0.2cm}

    \begin{subfigure}[b]{0.19\textwidth}
        \includegraphics[width=\textwidth]{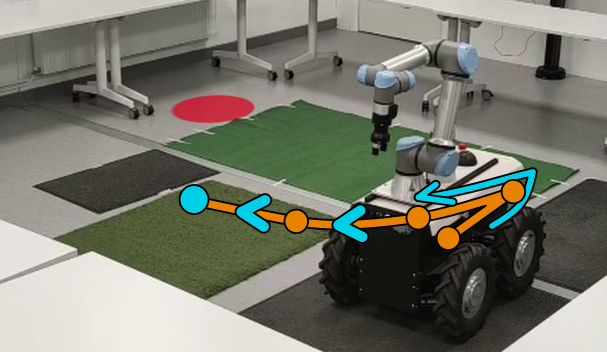}
    \end{subfigure}
    \hfill
    \begin{subfigure}[b]{0.19\textwidth}
        \includegraphics[width=\textwidth]{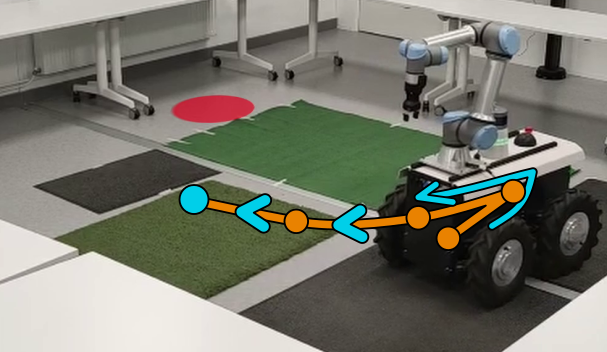}
    \end{subfigure}
    \hfill
    \begin{subfigure}[b]{0.19\textwidth}
        \includegraphics[width=\textwidth]{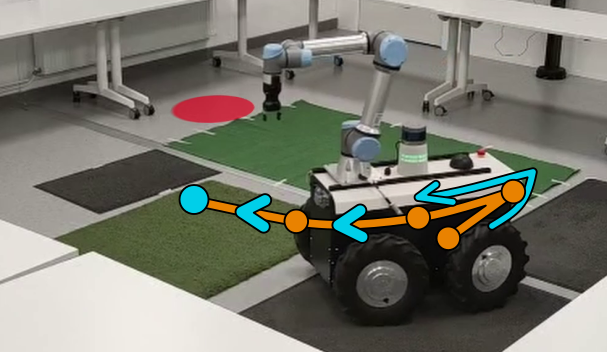}
    \end{subfigure}
    \hfill
    \begin{subfigure}[b]{0.19\textwidth}
        \includegraphics[width=\textwidth]{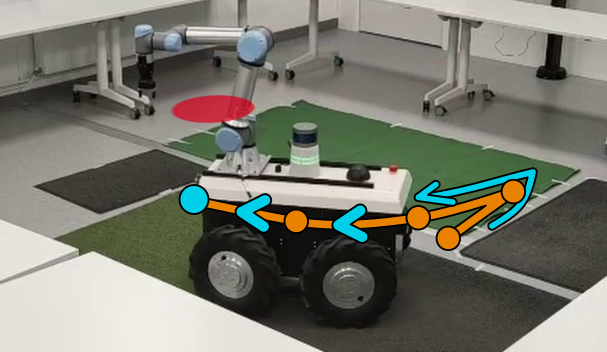}
    \end{subfigure}
    \hfill
    \begin{subfigure}[b]{0.19\textwidth}
        \includegraphics[width=\textwidth]{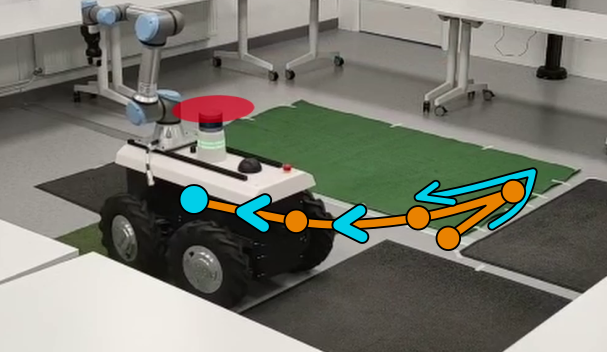}
    \end{subfigure}




    
    \vspace{-2pt}
    \caption{An example of a maneuver executed by ManeuverNet in both simulation and real-world settings is shown in the first two rows. Another example, showcasing multi-terrain performance, is presented in the final row of figures. The red dot represents $\boldsymbol{X}_d$.} \vspace{-0.5cm}
    
    \label{fig:bullet-vs-real}
\end{figure*}

\Update To assess the effectiveness of our specially designed reward functions for DASMRs (see Section~\ref{reward_new_section}), we performed a comparative study under different testing conditions. This study emphasizes the reward functions that best support precise and robust maneuver learning. \Update  Additionally we evaluated three reward functions adapted from the literature: \UpdateFinal 1) $\mathcal{R}_{\text{Exp}}$~\cite{SOUALHI2025RAS} based on the direction and distance to the target goal; \Done 2) $\mathcal{R}_{\text{HER}}$~\cite{Andrychowicz2017HER}  based on a sparse binary reward function; 3)  $\mathcal{R}_{\text{Car}}$ ~\cite{Lazzaroni2022APPLEPIES} combining the direction and distance to the target goal for a parking task. In this case, we omitted the collision term as it was unnecessary in our setup.





In this study, the agents \Done were tested in simulation using \Update PyBullet \Done under progressively challenging generalization settings, including: (i) using the same training seed and distance threshold $d_\text{th}$, (ii) using two unseen random seeds with $d_\text{th}= 15$~cm, (iii) evaluating all seeds with $d_\text{th}= 10$~cm. 
Setting (i) evaluated the agent under familiar conditions. In setting (ii), the sampled goal $\boldsymbol{X_d}$ is altered, which typically requires the robot to perform different maneuvers, thereby testing its ability to generalize to unseen spatial configurations. Setting (iii) evaluated the precision of the learned policy, demanding finer control near the goal. 


The corresponding results are summarized in Table~\ref{table:simulation_tests}. The results were generated over 100,000 time steps, with a new $\boldsymbol{X_d}$ sampled either when the agent reached the goal or when the episode was truncated after 300 time steps. As shown in Table~\ref{table:simulation_tests}, all state-of-the-art reward functions either failed entirely or converged to sub-optimal policies when applied to DASMRs. For example, $\mathcal{R}_\text{Car}$ and $\mathcal{R}_\text{Exp}$  performed poorly, suggesting that they are ill-suited to the maneuvering constraints inherent to DASMRs. Similarly, $\mathcal{R}_\text{HER}$ did not exceed a 55\% SR, indicating limited effectiveness.
While $\mathcal{R}_\text{Cl}$ achieved moderately better results (72\% and 62\% of SR), it struggled to generalize to unseen environments, highlighting its lack of robustness.

\UpdateFinal In contrast, the rewards $\mathcal{R}_\text{HS}$, $\mathcal{R}_\text{ES}$ and $\mathcal{R}_\text{Ch}$ demonstrated greater adaptability to complex maneuvering tasks. Among these, the reward $\mathcal{R}_\text{HS}$ performed the best overall, achieving at least 96\% of SR across all test scenarios with $d_\text{th} = 15$ cm and also superior performance in terms of the SPL metric, highlighting its efficiency and reliability. This improvement stems from the stronger weighting of the lateral ($y$) displacement in these rewards, which better captures the sideward maneuvering requirements inherent to DASMR. \Done


Additional results under challenging conditions, specifically continuous goal targeting without pose reinitialization, are available in our GitHub repository. These findings align with the main results and further showcase the flexibility and robustness of our reward functions.

\begin{table}[!tb]\vspace{0.1cm}
\captionsetup{font=scriptsize}
\caption{\Update Benchmarking results for an unseen seed and $d_{\text{th}} = 15$ cm, with the success rate (SR) in \%, the average error (AE) (the standard deviation $\sigma$)  in m, and the average SPL. \Done} \vspace{-0.4cm}
\scriptsize
\label{table:sim-to-sim}
\scriptsize
\begin{center}
\setlength{\tabcolsep}{4pt}
\begin{tabular}{l|l|*3c}
\toprule
{\textbf{Approach}} & \multirow{2}{*}{\textbf{Approach}} & \multicolumn{3}{c}{\textbf{Test Unseen: seed = 10}}\\
\cline{3-5}
{\textbf{Type}}&  & \textbf{SR $\uparrow$} & \textbf{AE($ \boldsymbol{\sigma}$) $\downarrow$} & \textbf{SPL $\uparrow$} \\                              
\hline
{\textbf{Analytical}}  & TEB~\cite{TEBref} & \textbf{100} & \textbf{0.05} (0.05) & \bf 0.79 \\
\hline
\hline
\multirow{3}{*} {\textbf{DRL}}&Standard DRL~\cite{DRLStandard}  & 45 & 0.25 (0.17) & 0.35 \\
\cline{2-5}
&FastRLap~\cite{Stachowicz2023CoRL} & 00 & 3.79 (2.11) & 0.00 \\
\cline{2-5}

&{ManeuverNet (ours)} & {\bf 85} & {\bf 0.15 (0.04)} & {\bf 0.70} \\

\bottomrule
\end{tabular} \vspace{-0.4cm}
\end{center}
\end{table}

\begin{table}[!t]
\captionsetup{font=scriptsize}
\caption{ManeuverNet simulation results for different reward functions on seen, unseen 1, and unseen 2 seeds, with the success rate (SR) in \%, the average error (AE) (the standard deviation $\sigma$)  in m, and the average SPL.} \vspace{-0.4cm}
\scriptsize
\label{table:simulation_tests}
\scriptsize
\begin{center}

\setlength{\tabcolsep}{1pt}
\begin{tabular}{l|c|*3c|*3c|*3c}
\toprule
\multirow{2}{*}{\textbf{Rew.}} & $\boldsymbol{d_\text{th}}$ & \multicolumn{3}{c|}{\textbf{Seen}} & \multicolumn{3}{c|}{\textbf{ Unseen 1}} & \multicolumn{3}{c}{\textbf{ Unseen 2}}\\
\cline{3-11}
& (cm) & \textbf{SR$\uparrow$} & \textbf{AE($ \boldsymbol{\sigma}$)$\downarrow$} & \textbf{SPL$\uparrow$} & \textbf{SR$\uparrow$} & \textbf{AE($ \boldsymbol{\sigma}$)$\downarrow$} & \textbf{SPL$\uparrow$} & \textbf{SR$\uparrow$} & \textbf{AE($ \boldsymbol{\sigma}$)$\downarrow$} & \textbf{SPL$\uparrow$}\\
\hline
\multirow{2}{*}{$\mathcal{R}_{\text{Car}}$} & 15 & 02 & 0.19(0.62) & 0.00 & 02 & 0.17(0.75) & 0.02 & 02 & 0.17(0.62) & 0.02\\
 & 10 & 02 & 0.19(0.63) & 0.00 & 02 & 0.18(0.76) & 0.02 & 02 & 0.17(0.63) & 0.02 \\
\hline
\multirow{2}{*}{$\mathcal{R}_{\text{Exp}}$} & 15 & 04 & 0.53(0.20) & 0.03 & 04 & 0.55(0.25) & 0.03 & 06 & 0.55(0.26) & 0.04\\
 & 10 & 02 & 0.53(0.24) & 0.01 & 02 & 0.56(0.24) & 0.01 & 00 & 0.57(0.24) & 0.00 \\
\hline
\multirow{2}{*}{{$\mathcal{R}_{\text{HER}}$}} & 15 & 55 & 0.25(0.21) & 0.47 & 50 & 0.28(0.25) & 0.40 & 55 & 0.30(0.26) & 0.45\\
 & 10 & 48 & 0.21(0.19) & 0.39 & 45 & 0.22(0.19) & 0.35 & 47 & 0.28(0.28) & 0.39 \\
\hline
\multirow{2}{*}{$\mathcal{R}_{\text{Cl}}$} & 15 & 72 & 0.27(0.25) & 0.64 & 58 & 0.38(0.34) & 0.50 & 60 & 0.34(0.29) & 0.54\\
 & 10 & 62 & 0.26(0.28) & 0.53 & 45 & 0.40(0.36) & 0.39 & 47 & 0.36(0.32) & 0.41 \\
\hline
\multirow{2}{*}{$\mathcal{R}_{\text{Ch}}$} & 15 & 92 & \textbf{0.15}(0.03) & 0.79 & 82 & \textbf{0.16}(0.06) & 0.70 & 88 & 0.17(0.12) & 0.77\\
 & 10 & 81 & 0.13(0.06) & 0.66 & 74 & 0.13(0.07) & 0.62 & 79 & 0.13(0.13) & 0.65 \\
\hline
\multirow{2}{*}{$\mathcal{R}_{\text{ES}}$} & 15 & \bf 97 & 0.16(0.07) & 0.81 & 96 & 0.17(0.16) & \bf 0.82 & 93 & 0.16(0.08) & 0.80\\
 & 10 & 88 & \textbf{0.12}(0.10) & 0.71 & \bf 88 & 0.12(0.11) & 0.71 & \bf 85 & \textbf{0.12}(0.09) & \bf 0.70 \\
\hline
\multirow{2}{*}{{$\mathcal{R}_{\text{HS}}$}} & 15 & \bf 97 & 0.16(0.22) & \bf 0.82 & \bf 97 & \textbf{0.16}(0.17) & \bf 0.82 & \bf 96 & \textbf{0.15}(0.03) & \bf 0.84\\
 & 10 & \bf 89 & 0.13(0.24) & \bf 0.73 & \bf 88 & \textbf{0.11}(0.04) & \bf 0.72 & \bf 85 & \textbf{0.12}(0.07) &  0.69 \\
                                
\bottomrule
\end{tabular} \vspace{-0.7cm}
\end{center}
\end{table}

\subsection{Real-World Results} \label{TEB_exp} 
We deployed ManeuverNet with $\mathcal{R}_\text{HS}$ on a real robot using zero-shot transfer, without any fine-tuning. The current state was estimated using the robot's sensors, which include an UM7 IMU, wheel encoders, and an RS-LiDAR-16 3D LiDAR. The robot was controlled via ROS within a real-world, square workspace measuring 4.2 meters per side, bounded by tables. Goal positions were randomly sampled within this area, and all evaluations were conducted using a fixed $d_\text{th} = 15$ cm. As illustrated in Fig.~\ref{fig:bullet-vs-real}, the DRL agent successfully controlled the robot with high precision in real-world conditions, despite the absence of any adaptation.

To further assess the generalization capabilities of ManeuverNet, we tested it on a variety of surface types, including vinyl, artificial grass, and carpet. An example is shown in Fig.~\ref{fig:bullet-vs-real}. Despite significant differences in ground friction and contact dynamics, the policy maintained consistent and robust performance, underscoring its independence from specific physical parameters of the simulation. Moreover, the robot was not always initialized at the center of the workspace, yet the agent reliably handled arbitrary initial positions and orientations. This demonstrates ManeuverNet's capacity to function effectively in unstructured and non-resettable environments.

\Update

To evaluate the effectiveness of ManeuverNet, we performed a comparative study with the TEB planner~\cite{TEBref}. 
TEB was configured to generate smooth trajectories toward the target positions. To ensure a fair comparison, we sampled the same six successful target positions for both ManeuverNet and the TEB planner. To ensure consistency in evaluation, the desired robot orientation for the TEB planner was matched to the final pose attained by the DRL agent.
 \Done 
The comparison in Table~\ref{table:TEB_vs_DRL} demonstrates that, in all targets tested, ManeuverNet consistently produced shorter paths, achieving up to a 90\% improvement in the SPL metric. \ToDo On average, ManeuverNet outperforms TEB by 40\% in SPL and reduces navigation time by 28\%.  \Update  Notably, similar trends have been observed in prior work~\cite{Arce2023}, which reported superior performance of DRL over TEB in the context of differential-drive robots.



Although both approaches successfully reached the six target positions, several practical limitations help explain the performance differences reported in Table~\ref{table:TEB_vs_DRL}. First, the TEB planner exhibits high sensitivity to the robot’s physical state: variations in dynamics, payload, or tire pressure often require fine-tuning and recalibration, which is impractical for real-world deployment. Second, the TEB planner includes a RANSAC-based algorithm for obstacle avoidance, which occasionally causes the robot to halt or oscillate between forward and backward motions, thereby impeding progress toward the goal. This behavior results in inefficient trajectories. Finally, the reliance of the TEB planner on a sequential planning/control loop introduces latency, reducing robot responsiveness during navigation. 





\begin{table}[tb]
\centering
\captionsetup{font=scriptsize}
\caption{\Update  Real-world comparison of SPL and [navigation time] (in seconds) across six target positions.  The final column reports the number of parameters requiring manual tuning for real-world deployment. $\boldsymbol{\Delta}^{+}$ denotes the performance improvement of ManeuverNet relative to the TEB planner.} \vspace{-0.2cm}
\scriptsize
\setlength{\tabcolsep}{4pt}
\scalebox{0.91}{
\begin{tabular}{l | c | c | c | c | c | c | c }
\toprule
\multirow{3}{*}{{\bf Approach}} & \multicolumn{6}{|c|}{\textbf{Target ID}} & \multirow{3}{*}{{\bf \#Param.}}\\
\cline{2-7}
& \textbf{1} &  \textbf{2} &  \textbf{3} &  \textbf{4} &  \textbf{5} &  \textbf{6} & \\
\cline{2-7}
& \multicolumn{6}{|c|}{\textbf{SPL$\uparrow$ [Time $\downarrow$] }} & \\ 
\cline{2-7}
\hline
TEB~\cite{TEBref} & 0.21 [9]  & 0.31 [4] & 0.63 [7]  & 0.93 [4]  & 0.28 [7]  & \textbf{0.37} [5]  & $>10$ \\
ManeuverNet & \textbf{0.40} [4]  & \textbf{0.32} [4] &  \textbf{1.00} [3] &  \textbf{0.99} [3] &  \textbf{0.41} [5] &  \textbf{0.37} [5] & \textbf{0}\\
\hline
\hline
$\boldsymbol{\Delta}^{+}$ ($\%$) & 90 [56] &  03 [00]  & 60 [57] &  06 [25] &  46 [29]  & 00 [00]  & $-$ \\

\bottomrule
\end{tabular}} \vspace{-0.5 cm}
\label{table:TEB_vs_DRL}
\end{table}



\Done


\Update
\subsection{Limitations}

Despite the strong performance of ManeuverNet, we have identified a few limitations. First, the current DRL approach does not account for obstacles. The agent is trained and evaluated in obstacle-free environments, which limits its applicability in cluttered or dynamic real-world scenarios where path planning and collision avoidance are critical. A promising solution to address this limitation is to integrate a higher-level planner such as A*, which could generate a global collision-free path to the target. This was successfully tested, \UpdateFinal as shown in the video: \url{https://youtu.be/3-aarbuEOSY}. Alternatively, we can incorporate obstacle avoidance in the learning process by extending the state space and adapting the reward function, as explored in our preliminary work~\cite{Daniel2025ManeuverNetWithObstacles}.\Done

Second, ManeuverNet focuses solely on reaching a desired position and does not explicitly handle the robot's final 2D pose. Nevertheless, our framework could be extended to handle precise final poses when needed by adapting the reward function to incorporate orientation constraints, similar to strategies used in goal-conditioned RL~\cite{LiuGCRL}.

\Done



\section{Conclusion}\label{sec:conclusion}

In this work, we proposed ManeuverNet, a DRL framework leveraging SAC and CrossQ, specifically tailored for the control of DASMRs. By designing novel reward functions that better take into account the maneuvering constraints of such robots, we addressed the limitations of existing state-of-the-art methods, which often fail to generalize or converge to sub-optimal policies in this scenario. ManeuverNet was evaluated extensively in simulation under various generalization settings. These included unseen goal distributions, stricter success thresholds, and non-reset scenarios. Across all settings, ManeuverNet consistently outperformed baseline approaches, improving success rate by at least 40\% while maintaining efficient trajectories. Furthermore, we demonstrated the framework's zero-shot transfer capabilities in real-world experiments, validating its robustness across diverse terrains and settings without requiring fine-tuning or expert demonstrations. \Update Moreover, when compared to the widely used TEB planner, ManeuverNet improved the maneuvering trajectory efficiency by up to 90\%. \Done  \Update For future work, we plan to extend our framework by integrating obstacle-aware planning and orientation control, further broadening the versatility and applicability in real-world scenarios. 
\Done  \vspace{-0.15cm}


\section*{Acknowledgment}
This work was supported by the French Government under the France 2030 program through the National Research Agency (ANR) grant reference ANR-24-PEAE-0002. It was also funded by the Nouvelle-Aquitaine Region through the MIRAE project. Author M. Aranda was supported through grant RYC2024-051408-I, funded by \mbox{MICIU/AEI/10.13039/501100011033} and by ESF+.

\bibliographystyle{IEEEtran}
\bibliography{biblio_iso4_abbreviations}

@book{SuttonMIT2018,
  author       = {Richard S. Sutton and Andrew G. Barto},
  title        = {{Reinforcement Learning - An Introduction}},
  publisher    = {{MIT} Press},
  year         = {2018}
}

@article{DanielRAL2024,
  author       = {M{\'{e}}lodie Daniel and
                  Aly Magassouba and
                  others},
  title        = {{Multi Actor-Critic {DDPG} for Robot Action Space Decomposition: {A}
                  Framework to Control Large 3D Deformation of Soft Linear Objects}},
  journal      = {{IEEE} Robot. Autom. Lett. (RA-L)},
  volume       = {9},
  number       = {2},
  pages        = {1318--1325},
  year         = {2024},
  doi          = {10.1109/LRA.2023.3342672}
}

@inproceedings{GaspardIROS2024,
  author       = {Cl{\'{e}}ment Gaspard and
                  Gr{\'{e}}goire Passault and
                  others},
  title        = {{FootstepNet: An Efficient Actor-Critic Method for Fast On-line Bipedal
                  Footstep Planning and Forecasting}},
  booktitle    = {{IEEE/RSJ} Int. Conf. Intell. Robots Syst. ({IROS})},
  pages        = {13749--13756},
  doi          = {10.1109/IROS58592.2024.10802320},
  year         = {2024}
}

@inproceedings{GaspardICRA2024,
  author       = {Cl{\'{e}}ment Gaspard and
                  Marc Duclusaud and
                  others},
  title        = {{{FRASA:} An End-to-End Reinforcement Learning Agent for Fall Recovery
                  and Stand Up of Humanoid Robots}},
    pages={15994-16000},
  booktitle = {IEEE Int. Conf. Robot. Autom. (ICRA)},
  year         = {2025}
  
}

@inproceedings{Haarnoja2018SAC,
  author       = {Tuomas Haarnoja and
                  Aurick Zhou and
                  others},
  title        = {{Soft Actor-Critic: Off-Policy Maximum Entropy Deep Reinforcement Learning
                  with a Stochastic Actor}},
  booktitle    = {PMLR Int. Conf. Mach. Learn. ({ICML})},
  volume       = {80},
  pages        = {1856--1865},
  year         = {2018}
}

@inproceedings{Fujimoto2018TD3,
  author       = {Scott Fujimoto and
                  Herke van Hoof and
                  David Meger},
  title        = {{Addressing Function Approximation Error in Actor-Critic Methods}},
  booktitle    = {PMLR Int. Conf. Mach. Learn. ({ICML})},
  volume       = {80},
  pages        = {1582--1591},
  year         = {2018}
}

@inproceedings{Bhatt2024CrossQ,
  author       = {Aditya Bhatt and
                  Daniel Palenicek and
                  others},
  title        = {{CrossQ: Batch Normalization in Deep Reinforcement Learning for Greater
                  Sample Efficiency and Simplicity}},
  booktitle    = {PMLR Int. Conf. Learn. Represent. ({ICLR})},
  year         = {2024}
}

@inproceedings{Zhao2020SimtoReal,
  author       = {Wenshuai Zhao and
                  Jorge Pe{\~{n}}a Queralta and
                  Tomi Westerlund},
  title        = {{{Sim-to-Real Transfer in Deep Reinforcement Learning for Robotics:
                  {A} Survey}}},
  booktitle    = {{IEEE} Symp. Ser. Comput. Intell. ({SSCI})},
  pages        = {737--744},
  year         = {2020},
  doi          = {10.1109/SSCI47803.2020.9308468}
}

@book{Siegwart2005,
  author = {Siegwart, R. and Nourbakhsh, I. R.},
  publisher = {{MIT} Press},
  title = {Introduction to Autonomous Mobile Robotics},
  year = {2005}
}

@article{Andrychowicz2017HER,
  title={{Hindsight Experience Replay}},
  author={Andrychowicz, Marcin and Wolski, Filip and others},
  journal={Adv. Neural Inf. Process. Syst.},
  volume={30},
  year={2017}
}

@InProceedings{Stachowicz2023CoRL,
  title = 	 {{FastRLAP: A System for Learning High-Speed Driving via Deep {RL} and Autonomous Practicing}},
  author =       {Stachowicz, Kyle and Shah, Dhruv and others},
  booktitle = 	 {PMLR Conf. Robot Learn. (CoRL)},
  pages = 	 {3100--3111},
  year = 	 {2023}
}

@article{SOUALHI2025RAS,
title = {{Leveraging Motion Perceptibility and Deep Reinforcement Learning for Visual Control of Nonholonomic Mobile Robots}},
journal = {Robot. Auton. Syst.},
volume = {189},
pages = {104920},
year = {2025},
doi = {10.1016/j.robot.2025.104920},
author = {Takieddine Soualhi and Nathan Crombez and others}
}

@article{Mehmood2021ITC,
author = {Mehmood, Atif and Shaikh, Inamulhasan and Ali, Ahsan},
year = {2021},
pages = {507-521},
number = {19},
title = {{Application of Deep Reinforcement Learning for Tracking Control of 3WD Omnidirectional Mobile Robot}},
volume = {50},
journal = {Inf. Technol. Control},
doi = {10.5755/j01.itc.50.3.25979}
}

@Article{Zhang2024AS,
AUTHOR = {Zhang, Yilin and Zeng, Jiayu and others},
TITLE = {{Dual-Layer Reinforcement Learning for Quadruped Robot Locomotion and Speed Control in Complex Environments}},
JOURNAL = {Appl. Sci.},
VOLUME = {14},
YEAR = {2024},
NUMBER = {19},
note={{Art. no. 8697}}}

@ARTICLE{Yu2010TRO,
  author={Yu, Wei and Chuy, Oscar Ylaya and others},
  journal={IEEE Trans. Robot. (T-RO)}, 
  title={{Analysis and Experimental Verification for Dynamic Modeling of A Skid-Steered Wheeled Vehicle}}, 
  year={2010},
  volume={26},
  number={2},
  pages={340-353},
  doi={10.1109/TRO.2010.2042540}}

@Article{Moreno2016Sensors,
AUTHOR = {Moreno, Javier and Clotet, Eduard and others},
title        = {{Design, Implementation and Validation of the Three-Wheel Holonomic Motion System of the Assistant Personal Robot {(APR)}}},
JOURNAL = {Sensors},
VOLUME = {16},
YEAR = {2016},
NUMBER = {10},
doi          = {10.3390/S16101658},
note = {{Art. no. 1658}}}

@Article{Honghu2022AS,
AUTHOR = {Xue, Honghu and Hein, Benedikt and others},
TITLE = {{Using Deep Reinforcement Learning with Automatic Curriculum Learning for Mapless Navigation in Intralogistics}},
JOURNAL = {Appl. Sci.},
VOLUME = {12},
YEAR = {2022},
NUMBER = {6},
note={{Art. no. 3153}}}

@article{Zhao2013AC,
author = {Jing-Shan Zhao and Xiang Liu and others},
title ={{Design of an Ackermann-type Steering Mechanism}},
journal = {Proc. Inst. Mech. Eng.},
volume = {227},
number = {11},
pages = {2549-2562},
year = {2013},
doi = {10.1177/0954406213475980},
}

@INPROCEEDINGS{Deremetz2017ECMR,
  author={Deremetz, Mathieu and Lenain, Roland and others},
  booktitle={Eur. Conf. Mob. Robot. (ECMR)}, 
  title={{Path Tracking of a Four-Wheel Steering Mobile Robot: A Robust Off-Road Parallel Steering Strategy}}, 
  year={2017},
  pages={1-7},
  doi={10.1109/ECMR.2017.8098670}}

@INPROCEEDINGS{Mihir2021,
  author={Patil, Mihir and Wehbe, Bilal and Valdenegro-Toro, Matias},
  booktitle={IEEE OCEANS}, 
  title={{Deep Reinforcement Learning for Continuous Docking Control of Autonomous Underwater Vehicles: A Benchmarking Study}}, 
  year={2021},
  pages={1-7},
  doi={10.23919/OCEANS44145.2021.9706000}}

@INPROCEEDINGS{Datar2024IROS,
  author={Datar, Aniket and Pan, Chenhui and others},
  booktitle={IEEE/RSJ Int. Conf. Intell. Robots Syst. (IROS)}, 
  title={{Terrain-Attentive Learning for Efficient 6-DoF Kinodynamic Modeling on Vertically Challenging Terrain}}, 
  year={2024},
  pages={5438-5443},
  doi={10.1109/IROS58592.2024.10801650}}

@inproceedings{Lazzaroni2022APPLEPIES,
  author       = {Luca Lazzaroni and
                  Francesco Bellotti and
                  others},
  title        = {{Deep Reinforcement Learning for Automated Car Parking}},
  booktitle    = {Springer Int. Conf. Appl. Electron. Pervading Ind. Environ. Soc. ({ApplePies})},
  volume       = {1036},
  pages        = {125--130},
  year         = {2022},
  doi          = {10.1007/978-3-031-30333-3\_16}
}

@article{Junzuo2021IOP,
doi = {10.1088/1742-6596/1883/1/012111},
year = {2021},
volume = {1883},
number = {1},
pages = {012111},
author = {Junzuo, Li and Qiang, Long},
title = {{An Automatic Parking Model Based on Deep Reinforcement Learning}},
journal = {IOP Publishing J. Phys. Conf. Ser.}
}

@article{double-ackermann-2,
  author       = {Redmond R. Shamshiri and
                  Alireza Azimi and
                  others},
  title        = {{Online Path Tracking With an Integrated H\({}_{\mbox{{\(\infty\)}}}\)
                  Robust Adaptive Controller for a Double-Ackermann Steering Robot for
                  Orchard Waypoint Navigation}},
  journal      = {Springer Int. J. Intell. Robot. Appl. (IJIRA)},
  volume       = {9},
  number       = {1},
  pages        = {257--277},
  year         = {2025},
  doi          = {10.1007/S41315-024-00379-2}
}

@article{SamsonTRO,
  author       = {Claude Samson},
  title        = {{Control of Chained Systems Application to Path Following and Time-Varying
                  Point-Stabilization of Mobile Robots}},
  journal      = {{IEEE} Trans. Robot. (T-RO)},
  volume       = {40},
  number       = {1},
  pages        = {64--77},
  year         = {1995},
  doi          = {10.1109/9.362899}
}

@article{Thuilot2009,
  author       = {Christophe Cariou and
                  Roland Lenain and
                  others},
  title        = {{Automatic Guidance of a Four-Wheel-Steering Mobile Robot for Accurate
                  Field Operations}},
  journal      = {J. Field Robot.},
  volume       = {26},
  number       = {6-7},
  pages        = {504--518},
  year         = {2009},
  doi          = {10.1002/ROB.20282}
}

@inproceedings{Hulttinen2020,
  author       = {Lionel Hulttinen and
                  Jouni Mattila},
  title        = {{Flow-Limited Path-Following Control of a Double Ackermann Steered
                  Hydraulic Mobile Manipulator}},
  booktitle    = {{IEEE/ASME} Int. Conf. Adv. Intell. Mechatron. ({AIM})},
  pages        = {625--630},
  year         = {2020},
  doi          = {10.1109/AIM43001.2020.9158987}
}

@article{AndersonSPL,
  author       = {Peter Anderson and
                  Angel X. Chang and
                  others},
  title        = {{On Evaluation of Embodied Navigation Agents}},
  journal      = {CoRR},
  volume       = {abs/1807.06757},
  year         = {2018}}

@inproceedings{LiuGCRL,
  author       = {Minghuan Liu and
                  Menghui Zhu and
                  Weinan Zhang},
  title        = {{Goal-Conditioned Reinforcement Learning: Problems and Solutions}},
  booktitle    = {Int. Jt. Conf. Artif. Intell. ({IJCAI})},
  pages        = {5502--5511},
  year         = {2022},
  doi          = {10.24963/IJCAI.2022/770}
}

@article{TEB_parameters_limitation,
  author       = {Bence Magyar and
                  Nikolaos Tsiogkas and
                  others},
  title        = {{Timed-Elastic Bands for Manipulation Motion Planning}},
  journal      = {{IEEE} Robot. Autom. Lett. (RA-L)},
  volume       = {4},
  number       = {4},
  pages        = {3513--3520},
  year         = {2019},
  doi          = {10.1109/LRA.2019.2927956}
}

@INPROCEEDINGS{TEBref,
  author={Rösmann, Christoph and Hoffmann, Frank and Bertram, Torsten},
  booktitle={IEEE/RSJ Int. Conf. Intell. Robots Syst. (IROS)}, 
  title={{Kinodynamic Trajectory Optimization and Control for Car-Like Robots}}, 
  year={2017},
  pages={5681-5686},
  doi={10.1109/IROS.2017.8206458}}

@ARTICLE{DRLStandard,
  author={Miranda, Victor R. F. and Neto, Armando A. and others},
  journal={IEEE Trans. Ind. Electron.}, 
  title={{Generalization in Deep Reinforcement Learning for Robotic Navigation by Reward Shaping}}, 
  year={2024},
  volume={71},
  number={6},
  pages={6013-6020},
  doi={10.1109/TIE.2023.3290244}}

@article{ZhangWTYWS25,
  author       = {Wei Zhang and
                  Shanze Wang and others},
  title        = {{{DRL-DCLP:} {A} Deep Reinforcement Learning-Based Dimension-Configurable Local Planner for Robot Navigation}},
  journal      = {{IEEE} IEEE Robot. Autom. Lett. (RA-L)},
  volume       = {10},
  number       = {4},
  pages        = {3636--3643},
  year         = {2025},
  doi          = {10.1109/LRA.2025.3544927},
}

@article {Arce2023,
    author = {D. Arce and 
             J. Solano and
             C. Beltrán},
    title = {{A Comparison Study between Traditional and Deep-Reinforcement-Learning-Based Algorithms for Indoor Autonomous Navigation in Dynamic Scenarios}},
    journal = {Sensors},
    volume       = {23},
    number       = {24},
    note      = {{Art. no. 9672}},
    year = {2023}
}

@inproceedings{Lillicrap2015DDPG,
  author={Lillicrap, Timothy P and Hunt, Jonathan J and others},
  title={{Continuous Control with Deep Reinforcement Learning}},
  year={2016},
  booktitle={PMLR Int. Conf. Learn. Represent. ({ICLR})}
}

@inproceedings{Daniel2025ManeuverNetWithObstacles,
  author       = {Kohio Deflesselle and
                  M{\'{e}}lodie Daniel and
                  others},
  title        = {{Towards Safe Maneuvering of Double-Ackermann-Steering Robots with
                  a Soft Actor-Critic Framework}},
  booktitle    = {SIAV-FM2L workshop organized within IEEE/RSJ Int. Conf. Intell. Robots Syst. (IROS)},
  year         = {2025}
}

\end{document}